\documentclass{article}

\usepackage{arxiv}

\usepackage[utf8]{inputenc} 
\usepackage[T1]{fontenc}    
\usepackage{hyperref}       
\usepackage{url}            
\usepackage{booktabs}       
\usepackage{amsfonts}       
\usepackage{nicefrac}       
\usepackage{microtype}      
\usepackage{lipsum}		
\usepackage{graphicx}
\usepackage{natbib}
\usepackage{doi}
\usepackage{graphicx}
\usepackage{subcaption}
\usepackage{amssymb}
\usepackage{subcaption}
\usepackage{graphicx}
\usepackage{subcaption}
\usepackage{csquotes}

\usepackage{dashrule}
\usepackage{eqnarray}
\usepackage{amsmath}

\usepackage{hyperref}

\title{A hybrid Deep Learning-based approach for Optimal Genotype by Environment Selection}

\author{ {\hspace{1mm}Zahra Khalilzadeh}\thanks{Use footnote for providing further
		information about author (webpage, alternative
		address)---\emph{not} for acknowledging funding agencies.} \\
	Department of Industrial and Manufacturing Systems Engineering\\
	Iowa State University\\
	Ames, IA 50010 \\
	\texttt{zahrakh@iastate.edu} \\
	\And
	Motahareh Kashanian \\
	Department of Industrial and Manufacturing Systems Engineering\\
	Iowa State University\\
	Ames, IA 50010 \\
	\texttt{motinaa@iastate.edu} \\
    \And
	Saeed Khaki \\
	Department of Industrial and Manufacturing Systems Engineering\\
	Iowa State University\\
	Ames, IA 50010 \\
	\texttt{skhaki@iastate.edu} \\
	\AND
	Lizhi Wang \\
	Department of Industrial and Manufacturing Systems Engineering \\
    Iowa State University\\
	Ames, IA 50010 \\
	\texttt{lzwang@iastate.edu} \\
}

\renewcommand{\shorttitle}{\textit{arXiv} Template}

\hypersetup{
pdftitle={A template for the arxiv style},
pdfsubject={q-bio.NC, q-bio.QM},
pdfauthor={David S.~Hippocampus, Elias D.~Striatum},
pdfkeywords={First keyword, Second keyword, More},
}

\begin{document}
\maketitle

\begin{abstract}
Accurately predicting crop yield is vital for enhancing agricultural breeding and ensuring crop production remains resilient in diverse climatic conditions. Integrating weather data throughout the crop growing season, especially for various genotypes, is crucial for these predictions. It represents a significant stride in comprehending how climate change affects a variety's adaptability. In the MLCAS2021 Crop Yield Prediction Challenge, the Third International Workshop on Machine Learning for Cyber-Agricultural Systems released a dataset for soybean hybrids consisting of 93,028 training performance records to predict yield for the 10,337 testing performance records. This dataset spanned 159 locations across 28 states in the U.S. and Canadian provinces over a 13-year period, from 2003 to 2015. It comprised details on 5838 distinct genotypes and daily weather data for a 214-day growing season, encompassing all possible location and year combinations. As one of the winning teams, we designed two novel convolutional neural network (CNN) architectures. The first proposed model combines CNN and fully-connected (FC) neural networks (CNN-DNN model). The second proposed model adds an LSTM layer at the end of the CNN part for the weather variables (CNN-LSTM-DNN model). The Generalized Ensemble Method (GEM) was then utilized to determine the optimal weights of the proposed CNN-based models to achieve higher accuracy than other baseline models. The GEM model we introduced demonstrated superior performance compared to all other baseline models employed in soybean yield prediction. It exhibited a lower RMSE ranging from 5.55\% to 39.88\%, a reduced MAE ranging from 5.34\% to 43.76\%, and a higher correlation coefficient ranging from 1.1\% to 10.79\% in comparison to the baseline models when evaluated on test data. The proposed CNN-DNN model was then employed to identify the best-performing genotypes for various locations and weather conditions, making yield predictions for all potential genotypes in each specific setting. The dataset provides unique genotype information on seeds, allowing investigation of the potential of planting genotypes based on weather variables. The proposed data-driven approach can be valuable for genotype selection in scenarios with limited testing years. We also performed a feature importance analysis utilizing Root Mean Square Error (RMSE) change to identify crucial predictors impacting our model's predictions. The location variable exhibited the highest RMSE change, emphasizing its pivotal role in predictions, followed by MG, year, and genotype, showcasing their significance during crop growth stages and across different years. In the weather category, MDNI and AP displayed higher RMSE changes, indicating their importance.    

\end{abstract}

\keywords{Convolutional Neural Network \and Genotype Selection \and Crop Yield Prediction \and Generalized Ensemble Method}

\section{Introduction} \label{Introduction}
The world's population is projected to reach almost 10 billion by 2050 \citep{united2017world}, and climate change is expected to have a significant impact on crop yields in the coming years. As a result, there is an urgent need to increase crop production in order to feed the growing population. The current global food production systems are facing several challenges such as the increasing frequency and severity of droughts, floods, heatwaves and increased pests and diseases, which are all associated with climate change \citep{kumar2016impact}. These challenges are likely to affect crop yields and food security, making it essential to develop new strategies to increase crop production.

One of the main strategies for increasing crop production is to develop climate-resilient crops through breeding programs. This involves selecting and crossbreeding plants that are better able to withstand the effects of climate change, such as drought or heat stress. Despite the focus on climate resilience in breeding programs, there is mounting evidence of the difficulties and challenges in creating crops capable of handling the effects of climate change. These challenges stem from the contradiction between the pressing need for breeding in response to climate change and the inadequate understanding of how genotype and environment interact with each other \citep{xiong2022climate}. Another approach is to use crop simulation models that integrate environmental information and tools into the breeding analysis process to tackle the effects of climate change and anticipate crop growth and yield under different climate scenarios \citep{de2020data, heslot2014integrating}. However, crop simulation models have limitations such as complexity, where the simulations may not be able to fully capture all the interactions of multiple factors such as genetics, environment and management practices, leading to inaccurate predictions. Additionally, data availability, validation, computational resources, the limitation of analyzing a limited number of genotypes, and simplification of reality in the models are other limitations of simulation crop modeling \citep{roberts2017comparing, hajjarpoor2022process}. To overcome the limitations of crop growth models, studies are emerging recently to utilize statistical methods as promising alternatives and complementary tools. Among these methods, Machine Learning (ML) is a practical statistical approach that has gained popularity due to advancements in big-data technologies and high-performance computing. ML algorithms can help farmers to increase crop production in response to climate change by providing capabilities such as crop yield prediction  \citep{shahhosseini2021corn,khaki2019crop}, climate change impact modeling  \cite{crane2018machine}, climate-smart crop breeding \cite{xu2022smart}, automation of farming equipment \cite{patil2016early}, market price prediction \cite{chen2021automated}, water management optimization \cite{lowe2022review}, disease and pest forecasting \citep{domingues2022machine}, and precision agriculture \cite{sharma2020machine}. These capabilities can help farmers to plan for and adapt to changing weather patterns, identify resilient crops, optimize crop management practices, and make better decisions to increase crop production. The challenge of effectively training ML algorithms is posed by the inconsistent spatial and temporal data regarding some of the production and management inputs, such as planting date, fertilizer application rate, and crop-specific data. This is a problem that needs to be addressed for efficient ML algorithm training.

Genotype $\times$ environment interaction is a challenging factor that limits the genotype selection for increased crop yields in unseen and new environments especially with the presence of global climate change. Plant breeders typically choose hybrids based on their desired traits and characteristics, such as yield, disease resistance, and quality. They first select parent plants with desirable traits and cross them to create a new hybrid. The new hybrids are then tested in various environments to determine their performance, finally the hybrids with the highest yield are selected \citep{bertan2007parental}. However, this approach can be extremely time-consuming and tedious due to the vast number of possible parent combinations that require testing \citep{khaki2020predicting}. This highlights the importance of having a data driven approach to select genotypes with the highest performance in response to climates as well as other environmental variables using limited years of field testing per genotype. For example, \cite{arzanipour2022evaluating}, suggests employing imputation methods to address the issue of incomplete data, particularly when certain crop types are not cultivated in every observed environment. This perspective views these absent data points not merely as traditional missing values but as potential opportunities for additional observations. In this study, we introduce a cutting-edge deep learning framework for predicting crop yields using environmental data and genotype information. The framework is designed to identify the most efficient genotype for each location and environment, by first forecasting crop yields based on the given weather conditions in each location for all available genotypes, and then selecting the optimal genotype with the highest yield in each specific location and environmental scenario. This strategy helps in enhancing policy and agricultural decision-making, optimizing production, and guaranteeing food security. To the best of our knowledge this is the first study to use a deep learning approach for optimal genotype $\times$ environment selection. 

Over the years, several machine learning algorithms have been employed for predicting performance of crops under different environmental conditions. These include Convolutional Neural Network (CNN) \citep{srivastava2022winter}, Long Short Term Memory (LSTM) networks \citep{shook2021crop}, Regression Tree (RT) \citep{veenadhari2011soybean}, Random Forests (RF), Support Vector Machine (SVM), K-Nearest Neighbor (KNN), XGBoost, Least Absolute Shrinkage and Selection Operator (LASSO) \citep{kang2020comparative}, and Deep Neural Network (DNN) \citep{khaki2019crop}. In time series prediction tasks, deep neural networks have proven to be robust to inputs with noise and possess the ability to model complex non-linear functions \cite{dorffner1996neural}. By utilizing deep learning models, it becomes possible to tackle complex data, as these models can effectively learn the non-linear relationships between the multivariate input data, which includes weather variables, maturity group/cluster information, genotype information, and the predicted yield. 


Our proposed hybrid CNN-LSTM model consists of convolutional neural networks (CNNs) and Long Short-Term Memory (LSTM). CNNs can handle data in multiple array formats, such as one-dimensional data like signals and sequences, two-dimensional data such as images, and three-dimensional data like videos. A typical CNN model consists of a series of convolutional and pooling layers, followed by a few fully connected (FC) layers. There are several design parameters that can be adjusted in CNNs, including the number of filters, filter size, type of padding, and stride. Filters are weight matrices used to process the input data during convolution. Padding involves adding zeroes to the input data to maintain its dimensional structure, while the stride refers to the distance by which the filter is moved during processing \citep{albawi2017understanding}. Recurrent Neural Networks (RNNs) are a type of deep learning model designed for handling sequential data. The key advantage of RNNs is their ability to capture time dependencies in sequential data due to their memory mechanism, allowing them to use information from previous time steps in future predictions \citep{sherstinsky2020fundamentals, lipton2015critical}. Long Short-Term Memory (LSTM) networks are a specialized type of Recurrent Neural Network (RNN) that address the issue of vanishing gradients in traditional RNNs \citep{hochreiter1997long, sherstinsky2020fundamentals}. LSTMs are particularly beneficial for capturing long-term dependencies in sequential data, and they maintain information for longer periods of time compared to traditional RNNs \citep{hochreiter1996lstm}. These characteristics make LSTMs highly effective for handling data with complex temporal structures, such as speech and video \citep{xie2019attention, li2019residual}. Furthermore, LSTMs have been successfully utilized in multivariate time series prediction problems \citep{shook2021crop, sun2019county, gangopadhyay2018temporal}, and they are flexible and handle varying length inputs, making them suitable for processing sequential data with different lengths \citep{sutskever2014sequence}.

Crop yield prediction has been more recently improved by the application of deep learning methods. \cite{khaki2019crop} utilized deep neural networks to predict corn yield for various maize hybrids using environmental data and genotype information. Their study involved designing a deep neural network model that could forecast corn yield across 2,247 locations from 2008 to 2016. With regards to the accuracy of their predictions, the model they developed outperformed others such as Lasso, shallow neural networks, and regression trees, exhibiting a root-mean-square-error (RMSE) of 12$\%$ of the average yield when using weather data that had been predicted, and an RMSE of 11$\%$ of the average yield when using perfect weather data. Environmental data including weather and soil information and management practices were used as inputs to the CNN-RNN model developed by \cite{khaki2020cnn} for corn and soybean yield prediction across the entire Corn Belt in the U.S. for the years 2016, 2017, and 2018. Their proposed CNN-RNN model outperformed other models tested including random forest, deep fully connected neural networks, and LASSO, achieving a notable improvement with an RMSE of 9$\%$ and 8$\%$ of corn and soybean average yields, respectively. They also employed a guided backpropagation technique to select features and enhance the model's interpretability. Similarly, \cite{sun2019county} adopted a comparable strategy, utilizing a CNN-LSTM model to predict county-level soybean yields in the U.S. using satellite imagery, climate data, and other socioeconomic factors. Their results show that the CNN-LSTM model can capture the spatiotemporal dynamics of soybean growth and outperform other models in terms of accuracy and computational efficiency. \cite{oikonomidis2022hybrid} utilized a publicly available soybean dataset, incorporating weather and soil parameters to develop several hybrid deep learning-based models for crop yield prediction. Comparing their models with the XGBoost algorithm, the authors found that their hybrid CNN-DNN model outperformed the other models with an impressive RMSE of 0.266, MSE of 0.071, and MAE of 0.199. However, none of these studies have addressed the issue of determining which crop genotype to plant based on the given weather conditions. The dataset, which was developed, prepared, and cleaned by \cite{shook2021crop}, provided us with unique genotype information on seeds, allowing us to investigate the potential of planting genotypes based on weather variables. 
Our proposed data-driven approach can be particularly valuable for selecting optimal genotypes when there are limited years of testing available. This is because the traditional approach of selecting the best genotypes based on a small number of years of field trials can be unreliable due to variations in weather and other environmental factors. By leveraging large datasets with genotype and weather information, it becomes possible to develop more accurate models that can predict the performance of different genotypes in various weather conditions. This can ultimately lead to the identification of genotypes that are both high-yielding and adaptable to different environments. Given that land for agriculture is limited, such data-driven approaches can help improve the productivity of crops per acre, as well as the quality and productivity of food crops through plant breeding.

This study has three main objectives. Firstly, it proposes two novel convolutional neural network (CNN) architectures that incorporate a 1-D convolution operation and a long short-term memory (LSTM) layer. To achieve higher accuracy than other baseline models, the Generalized Ensemble Method (GEM) is utilized to determine the optimal weights of the proposed CNN-based models. Secondly, the proposed CNN-DNN model is utilized to select optimal genotypes for each location and weather condition. This is achieved by predicting the yield for all possible genotypes in each specific location and environmental scenario. Lastly, the study assesses the impact of location, maturity group (MG), genotype, and weather variables on prediction outcomes, investigating critical time periods for weather variables in yield predictions throughout the growing season of 30 weeks. Through these objectives, this study demonstrates the value of using data-driven approaches in plant breeding and crop productivity research.

The structure of this paper is as follows. Section \ref{Data} introduces the dataset used in this study. In Section \ref{Method}, we propose a methodology for crop yield prediction and optimal genotype selection using two CNN-based architectures with a 1-D convolution operation and LSTM layer, as well as the Generalized Ensemble Method (GEM) to find optimal model weights. This section also includes implementation details of the models used in this research, along with the design of experiments. Section \ref{Results} presents the experimental results, followed by an analysis of the findings in Section \ref{Analysis}. Finally, in Section \ref{Conclusion}, we conclude the paper by discussing the contributions of this work and highlighting potential avenues for future research.

\section{Data} \label{Data}
In this paper, the data analyzed was taken from the MLCAS2021 Crop Yield Prediction Challenge (\cite{mlcas2021challenge}) and consisted of 93,028 training and 10,337 testing performance records from 159 locations across 28 states in the U.S. and Canadian provinces, over 13 years (2003 to 2015). The data included information on 5838 unique genotypes and daily weather data for a 214-day growing season, covering all location and year combinations. This data was prepared and cleaned by \cite{shook2021crop}. The unique characteristic of this dataset is that it enables us to capture the biological interactions complexity, and temporal correlations of weather variables, as it provides both daily weather variables during the growing season for different locations and genotype data. The dataset included a set of variables for each performance record, which are as follows: 

$\bullet$ Weather: Every performance record in the dataset included a multivariate time-series data for 214 days, which represent the crop growing season between April 1\textsuperscript{st} and October 31\textsuperscript{st}. Each day in the record contained seven weather variables, including average direct normal irradiance (ADNI), average precipitation (AP), average relative humidity (ARH), maximum direct normal irradiance (MDNI), maximum surface temperature (MaxSur), minimum surface temperature (MinSur), and average surface temperature (AvgSur). Records with the same location and yield year share the same set of weather variables.

$\bullet$ Maturity group (MG): Dataset included 10 maturity groups corresponding to different regions. 

$\bullet$ Genotype IDs: The dataset contained 5838 distinct genotypes, which were further clustered into 20 groups using the K-means clustering technique as described in \cite{shook2021crop}. The resulting hard clustering approach allowed us to obtain a unique cluster ID for each of the 5839 genotypes in the dataset.

$\bullet$ State: The state information was provided for each performance record, indicating the specific state that the record corresponds to. The data covers 28 U.S. states and Canadian provinces in total.

$\bullet$ Location ID: For each performance record, the dataset included the corresponding location ID, indicating the unique identifier for the location associated with the record. The data was collected from a total of 159 locations.

$\bullet$ Year: The performance record dataset contained information on the year when the yield was recorded, ranging from 2003 to 2015. 

$\bullet$ Yield: The yield performance dataset included the observed average yield of soybean in bushels per acre across the locations in 28 U.S. states and Canadian provinces, between the years 2003 and 2015. 

\begin{figure}[ht]
  \centering
  \begin{subfigure}[b]{0.4\textwidth}
    \centering
    \includegraphics[width=\textwidth]{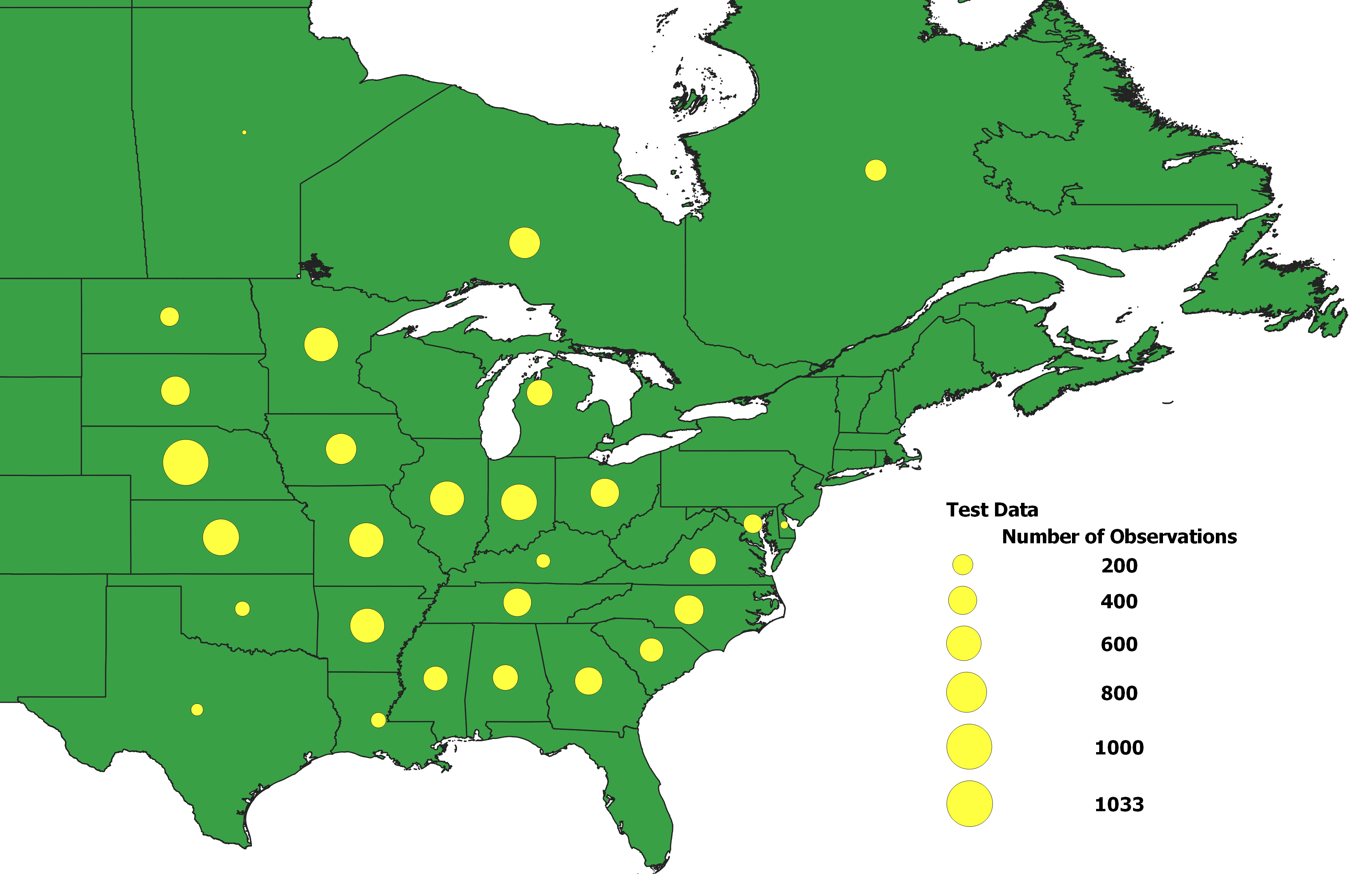}
    \caption{}
    \label{fig:subfig_a}
  \end{subfigure}
  \hfill
  \begin{subfigure}[b]{0.4\textwidth}
    \centering
    \includegraphics[width=\textwidth]{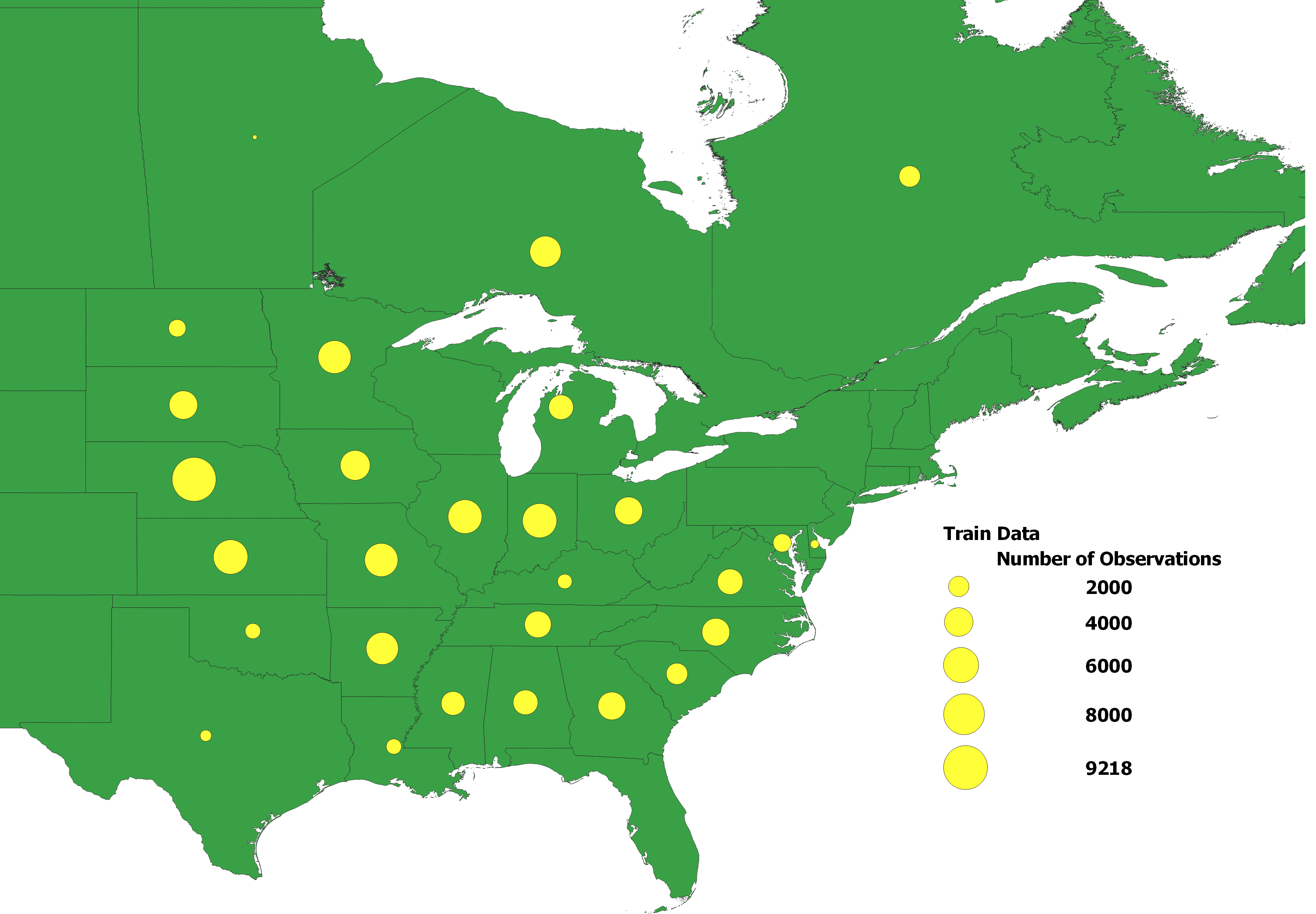}
    \caption{}
    \label{fig:subfig_b}
  \end{subfigure}
  \caption{The distribution of performance records across 28 U.S. states and Canadian provinces in the test (a) and train (b) datasets. The size of each yellow dot corresponds to the size of the dataset for the corresponding state\textbackslash province.}
  \label{datadistribution}
\end{figure}

The goal of the 2021 MLCAS Crop Yield Prediction Challenge was to predict soybean yield for the test data consisting of 10,337 performance records including observations from all years and locations. Our proposed GEM model achieved an impressive RMSE of 5.95 and MAE of 4.47 on the test set, earning us third place in the competition. In this paper we did one step further and used our proposed CNN-DNN model to  select the top 10 optimal genotypes with the highest yields in each specific location and environmental scenario. Since the ground truth response variables for the test data were not released after the competition, we only used the training dataset for yield prediction and further analysis. Specifically, we used the training dataset to select the top 10 optimal genotypes with the highest yields in each specific location and environmental scenario. Figure \ref{datadistribution} displays the distribution of performance records across 28 U.S. states and Canadian provinces in the test and train datasets. The size of each yellow dot corresponds to the size of the dataset for the corresponding state\textbackslash province. Table \ref{Summarystatistics} provides an overview of the summary statistics for both the dependent variable, soybean yield, and all independent variables used in the study(only training dataset).



\begin{table}[h]
\centering
\caption{Summary statistics of soybean yield data. The unit of yield is bushels per acre.}
\label{tab:yield-summary}
\begin{tabular}{l c}
\hline
\textbf{Summary Statistics} & \textbf{Value} \\
\hline
Total number of locations & 159 \\ 
Year range & 2003-2015 \\
Mean yield & 50.66 \\
Standard deviation of yield & 15.95 \\
25th percentile of yield  & 39.8 \\
Median yield & 50.60 \\
75th percentile of yield & 61.40 \\
Minimum yield & 0.4 \\
Maximum yield & 112.40 \\
Number of weather components & 7 \\
Number of maturity groups & 10 \\
Number of genotype IDs & 5838 \\
Number of observations & 93,028 \\
\hline
\label{Summarystatistics}
\end{tabular}
\end{table}


\section{Method} \label{Method}
\subsection{Data Preprocessing} \label{DataPreprocessing}
The pre-processing tasks were conducted to ensure the data is in a useful and efficient format for fitting machine learning models.  One of the main tasks involved one-hot encoding the categorical variables, which included maturity group, year, location IDs, and genotype IDs. For the genotype data we tried both genotype clusters and the unique genotypes. The results demonstrated a significant improvement when the genotype IDs were included with other variables. In one-hot encoding, each unique value of each categorical variable is represented as a new binary feature in a new column. This means that for every observation, a value of 1 is assigned to the feature that corresponds to its original category, while all other features are set to 0. This technique results in a new binary feature being created for each possible category, allowing for more accurate modeling and prediction.

To reduce the complexity of the daily weather data and make it more suitable for analysis, we aggregated the feature values by taking the average and downsampling the data to a 4-day level.  As a result of this downsampling and feature aggregation, we were able to reduce the number of model parameters significantly, with a dimension reduction ratio of 214:53. Reducing the daily weather data to a weekly level through downsampling has been commonly utilized in yield prediction studies to address the issue of excessive granularity in the data. This practice has been validated in prior research studies \citep{khaki2019crop, shook2021crop, srivastava2022winter}. 

Given the diverse range of values and varying scales of weather variables, it is important to avoid bias that may arise from a single feature. To address this, we applied the z-score normalization technique (Equation \ref{normalization}) to standardize all weather variable values. This technique rescales all weather variables to conform to a standard normal distribution, preventing any unintended bias on the results. In addition to mitigating bias, standardizing the weather variable values also improves the numerical robustness of the models and accelerates the training speed. 
\begin{equation} \label{normalization}
W_{i,j} = \frac{w_{i,j} - \bar{w}_j}{\sigma_j} 
\end{equation}

Where $W_{i,j}$ is the standardized value of the $i$th observation of the $j$th weather variable ($j$ ranges from 1 to $K$, where $K$ represents the total number of weather variables, which in this case is 371 (7 variables * 53 time periods)), $w_{i,j}$ is the original value of the $i$th observation of the $j$th weather variable, $\bar{w}_j$ is the mean of the $j$th weather variable, and $\sigma_j$ is the standard deviation of the $j$th weather variable. The formula rescales each variable to have a mean of 0 and a standard deviation of 1. 

\subsection{Model development} \label{Modeldevelopment}
In this section, we introduce two proposed models, CNN-DNN and CNN-LSTM-DNN, for predicting crop yield using location, MG,
genotype, and weather data. These models are designed to handle the temporal features of weather data, which play a crucial role in crop yield prediction. CNN-DNN is a combination of Convolutional Neural Networks (CNNs) and Deep Neural Networks (DNNs), while CNN-LSTM-DNN is a combination of CNNs, Long Short-Term Memory (LSTM) Networks, and DNNs. Both models are trained and evaluated using the same dataset. 

To improve the accuracy of our yield predictions, we propose using a Generalized Ensemble Method (GEM) that combines the predictions of both models. This approach allows us to leverage the strengths of each model and obtain better root mean squared error (RMSE) values than either model alone. In the following subsections, we describe the architecture and training procedures for the CNN-DNN and CNN-LSTM-DNN models, as well as the implementation of the GEM method for the yield prediction.

\subsubsection{Proposed CNN-DNN Model} \label{CNN-DNNModel}
The first proposed model architecture combines convolutional neural networks (CNN) and fully$-$connected (FC) neural networks. The weather variables measured throughout the growing season are taken as input in the convolutional neural network part of the model, which captures their temporal dependencies, and linear and nonlinear effects through 1$-$dimensional convolution operations. The CNN part of the model takes in the seven weather variables separately and concatenates their corresponding output for capturing their high$-$level features. The data for genotype, maturity group, location, and year (input \textunderscore others) are fed into a fully$-$connected neural network with one layer. The high$-$level features from the CNN are then combined with the output of the fully$-$connected neural network for input \textunderscore others data. The combined features are then processed through three additional FC layers before yielding the final prediction of the soybean yield. Moreover, to prevent overfitting, three dropout layers with dropout ratios of 0.5, 0.7, and 0.2 are respectively added to the fully connected layer after the CNN layer, at the end of the fully connected layer for input \textunderscore others data, and at the final layer of the model. The proposed modeling architecture is designed to capture the complex interactions between weather data, genotype IDs, maturity groups, year, and location IDs for an accurate yield prediction and is illustrated in Fig \ref{fig:a}.  



\subsubsection{Proposed CNN-LSTM-DNN Model} \label{CNN-LSTM-DNNModel}
\begin{figure}[ht]
\centering
\begin{subfigure}{\textwidth}
  \centering
  \includegraphics[width=0.8\linewidth]{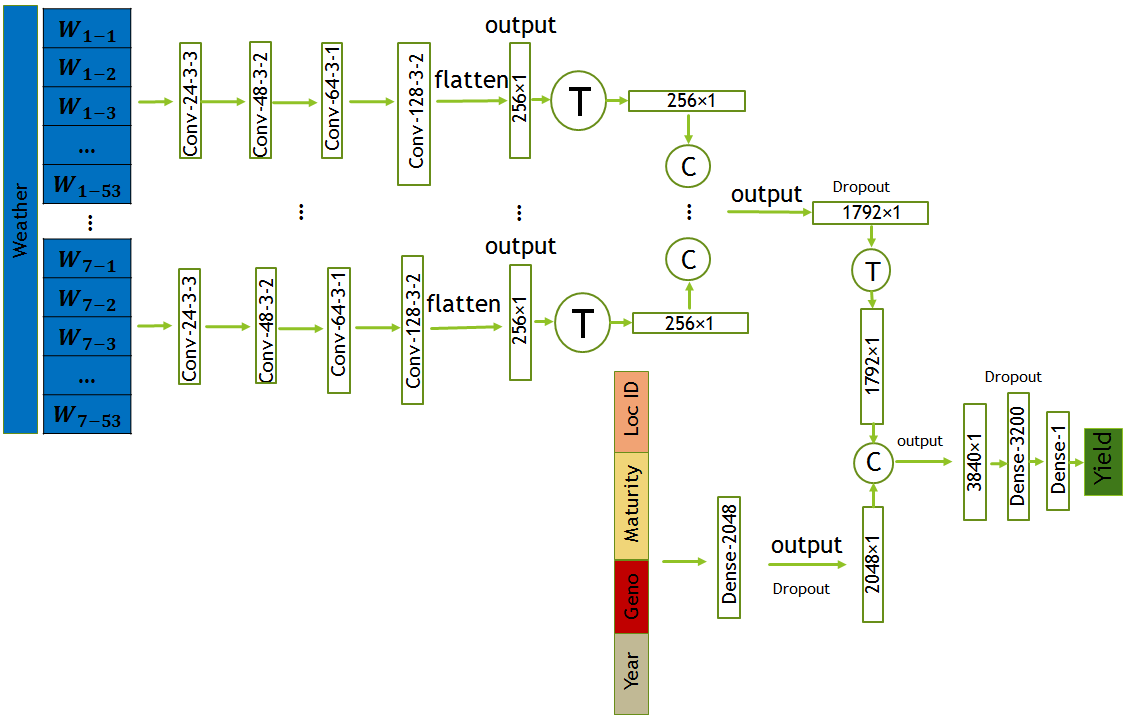}
  \caption{Proposed CNN-DNN Model}
  \label{fig:a}
\end{subfigure}
\begin{subfigure}{\textwidth}
  \centering
  \includegraphics[width=0.8\linewidth]{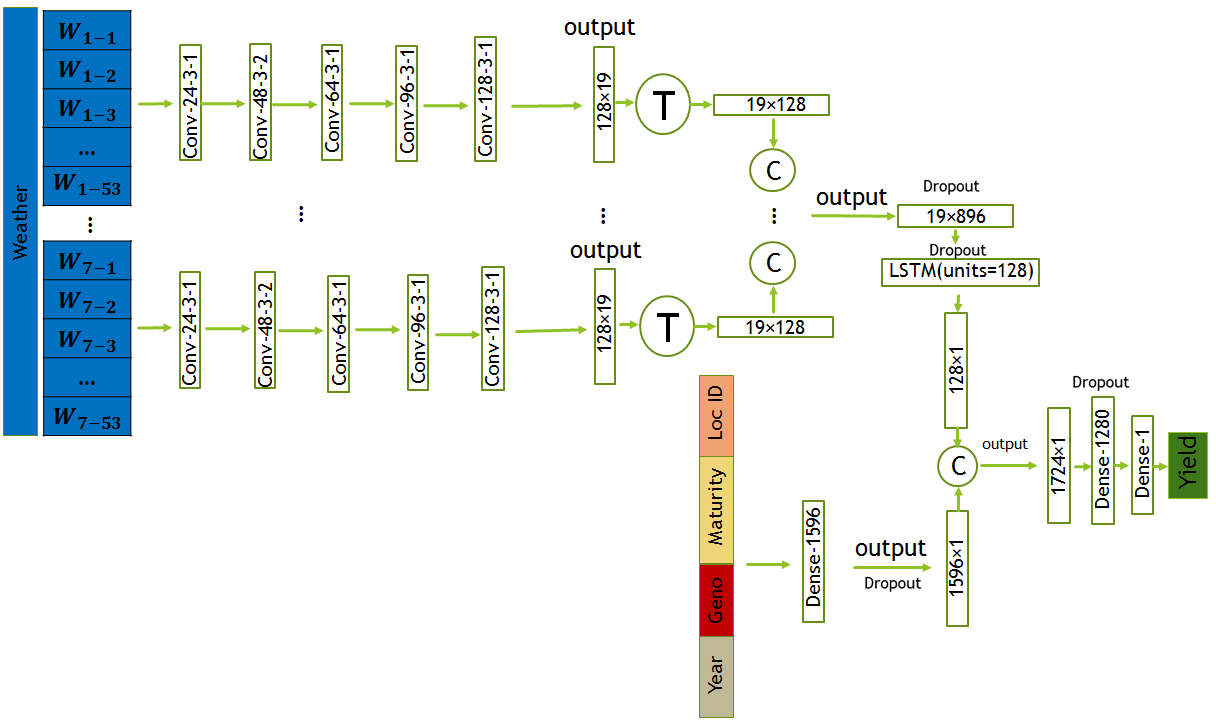}
  \caption{Proposed CNN-LSTM-DNN Model}
  \label{fig:b}
\end{subfigure}
\caption{The CNN architectures proposed in this study includes convolutional, and fully connected layers denoted by Conv, and Dense respectively. The parameters of the convolutional layers are presented in the form of \enquote{convolution type— number of filters—kernel size—stride size}. For all layers, \enquote{valid} padding was employed. Matrix concatenations are indicated by \textcircled{C}, while the symbol \textcircled{T} is used to indicate matrix transpose. Rectified Linear Unit (ReLU) was chosen as the activation function for all networks, except for the fully connected layers in input \textunderscore other data, which did not have an activation function.  }
\label{fig:ab}
\end{figure}

The second proposed model shares the same architecture as the first one, with the addition of an LSTM layer at the end of the CNN part for the weather variables. Specifically, the output of the CNN part is passed to an LSTM layer consisting of 128 units. The resulting output is then combined with the output of the fully connected layer for the input \textunderscore others data. This model architecture is designed to further capture the temporal dependencies and nonlinear effects of the weather variables, in addition to the high-level features extracted by the CNN part. In addition to the architecture described above, dropout layers were utilized to prevent overfitting. Specifically, four dropout layers with dropout ratios of 0.5, 0.5, 0.7, and 0.2 were respectively inserted after the CNN layer, at the LSTM layer, at the end of the fully connected layer for input \textunderscore other data, and at the final layer of the model. The complete modeling architecture is illustrated in Fig \ref{fig:b}.


\subsubsection{Generalized Ensemble Method (GEM)} \label{GeneralizedEnsembleMethod}
The Generalized Ensemble Method (GEM) is an advanced technique for creating a regression ensemble that combines the strengths of multiple base estimators. The method was first proposed by \cite{perrone1995networks} in the context of artificial neural networks. The main goal of GEM is to find the optimal weights of the base models that minimize the error metric, such as mean squared error (MSE) or root mean squared error (RMSE). To prepare the data for model training and evaluation, we randomly partitioned the dataset into a training set containing 80 \% of the data (74,422 samples), a validation set including 10 \% of the data (9,303 samples), and a test set containing the remaining 10 \% of the data (9,303 samples). We selected the best performing model on the validation set and leveraged the following optimization approach to create an ensemble of models that further improved the prediction accuracy. The problem can be stated as a nonlinear convex optimization problem, where the objective is to minimize the sum of squared errors between the true values ($y_{i}$) and the predicted values ($\hat{y}_{ij}$) of all observations (i=1…,n) by the k base models (j=1…,k). The validation set was used to optimize the ensemble weights.


\begin{eqnarray}
\min_{w_j} & \frac{1}{n} \sum_{i=1}^{n} (y_i - \sum_{j=1}^{k} w_j  \hat{y}_{ij}) ^ 2 & \label{GEMModel}
\end{eqnarray}

The problem is subject to two constraints: the weights of all base models should be non-negative ($w_j \geq 0$) and sum up to one ($\sum_{j=1}^{k} (w_j) = 1 $). Here, $w_j$ represents the weight assigned to base model j.

\subsection{Optimal Genotype by Environment Selection} \label{OptimalGenotypeby EnvrionmentSelection}
To predict the yield for all possible genotypes in each specific location and environmental scenario, we employed the CNN-DNN model. To simplify the results, we excluded the maturity group feature and retrained the proposed model. This was necessary because different types of maturity groups were utilized in each location. The aim of this approach is to showcase the best genotypes that could be cultivated in each specific location and weather (year) scenario.  

\subsection{Design of Experiments} \label{DesignofExperiments}

Since the ground truth response variables for the test data were not released after the competition, we solely relied on the the training dataset, which consisted of 93,028 observations, to train and test our proposed DL models and other ML models. The data preprocessing resulted in 6391 column features (6020 features after one-hot encoding maturity group, year, location IDs, and genotype IDs, and 371 (53*7) features after downsampling the weather data to a 4-day level). In order to make a comprehensive comparison, we incorporated three additional commonly used prediction models: random forest (RF) \citep{breiman2001random}, extreme gradient boosting (XGBoost) \citep{chen2016xgboost}, and least absolute shrinkage and selection operator (LASSO) \citep{tibshirani1996regression}. Further details on the implementation of these models are outlined below. 

\textbullet RF is an ensemble learning algorithm in machine learning. It works by combining multiple decision trees and using bagging to create a more accurate model. Bagging involves randomly sampling the dataset multiple times with replacement, creating different subsets of data for each tree to learn from. Each tree is trained on a different subset of the data, and their predictions are combined to make the final prediction. After experimenting with various numbers of trees in the RF model, we discovered that 550 trees produced the most accurate predictions. Furthermore, increasing the number of trees did not improve the accuracy but significantly increased the training time. We also examined different numbers of maximum tree depths and observed that a maximum depth of 55 generated the most precise predictions. Altering the maximum depth of the trees had a significant impact on the prediction accuracy, with an increase resulting in overfitting and a decrease leading to decreased prediction accuracy.

\textbullet XGBOOST is a popular machine learning algorithm known for its speed and accuracy in solving regression and classification problems. It is based on the concept of boosting, where weak learners are combined to form a strong learner. XGBOOST is an optimized version of gradient boosting, and it uses a tree-based model. It has several advantages, such as handling missing values, feature importance ranking, and regularization to prevent overfitting. To optimize the XGBOOST model for predicting soybean yield, we explored different hyperparameters ranges for max depth and subsample. After training and validating multiple models with different combinations of hyperparameters, we found that a max depth of 13 and a subsample of 0.7 provided the best results in terms of RMSE and MAE.

\textbullet LASSO is a linear regression technique used to analyze data with a high number of features. It uses regularization to constrain the coefficient estimates towards zero, which results in simpler models and reduces the risk of overfitting. The LASSO model adds a penalty term to the sum of the squared residuals, where the penalty is proportional to the absolute value of the coefficients. The optimization algorithm tries to minimize this penalty term along with the sum of the squared residuals. The alpha parameter in sklearn's Lasso function controls the strength of the L1 penalty on the coefficients, which is the same as the L1 term in the LASSO model. A higher alpha value will result in more coefficients being forced to zero, leading to a simpler and more interpretable model. In our study, we tried a range of alpha values, including [0.0001, 0.001, 0.01, 0.1, 1, 10, 100], and found that alpha=0.0001 provided the best result.


We maintained the same randomly partitioned dataset, which was used to train the ensemble models, consisting of a training set with 80$\%$ of the data (74,422 samples), a validation set with 10$\%$ of the data (9,303 samples), and a test set with the remaining 10$\%$ of the data (9,303 samples), for both hyperparameter tuning and model evaluation. Multiple models were trained using various hyperparameter values and their performance was evaluated on the validation set. The hyperparameter values that resulted in the best performance on the validation set were selected, and the corresponding model was evaluated on the test set to estimate its generalization performance. The range of hyperparameter values that we tested was selected based on our domain knowledge. Table \ref{MLhyperparams} shows the tested hyperparameters along with the best estimates obtained for the baseline models.

\begin{table}[h]
\centering
\caption{Hyperparameters of the baseline machine learning models employed to predict soybean yield.}
\label{MLhyperparams}
\begin{tabular}{l c c}
   \hline
        \textbf{Model} & \textbf{Parameters} & \textbf{Best Parameter} \\
        \hline
        Random Forest & Number of estimators & 550 \\
         & Max. feature numbers & Sqrt \\
         & Max. depth & 55 \\
         & Min. samples split & 5 \\
         & Min. samples leaf & 1 \\
         & Bootstrap & FALSE \\
        \hline
        XGBoost & Max. depth & 13 \\
         & Objective & [reg:squared error] \\
         & regularization alpha &  0.0001 \\
         & Min. child weight & 5 \\
         & Gamma &  0.05 \\
         & Learning rate & 0.09 \\
         & Booster & Gbtree \\
         & Subsample & 0.7 \\
         & Column sample by tree & 0.9 \\
         \hline
        Lasso regression & alpha & 0.0001\\
         \hline
\end{tabular}
\end{table}


The architecture and hyperparameters of the CNN-DNN and CNN-LSTM-DNN models are described in figures \ref{fig:a} and \ref{fig:b}, respectively. We trained the proposed models using the Adam optimizer with a scheduled learning rate of 0.0004, which decayed exponentially with a rate of 0.96 every 2500 steps. The models were trained for 800,000 iterations with a batch size of 48. Rectified Linear Unit (ReLU) was chosen as the activation function for all networks, except for the fully connected layers for input \textunderscore other data, which did not have an activation function. 



\subsection{Model Evaluation} \label{Modelevaluation}
In this study, we evaluated the performance of our prediction models using two widely used metrics: mean absolute error (MAE) [Eq. \ref{MAE}] and root mean square error (RMSE) [Eq. \ref{RMSE}]. Both of these metrics provide a measure of the distance between the predicted and actual values of the target variable. Specifically, MAE represents the average absolute difference between the predicted and actual values, while RMSE represents the square root of the average of the squared differences between the predicted and actual values. By using both of these metrics, we were able to assess the accuracy of our models and compare their performance against each other. We also reported the results of correlation coefficient (r) [Eq. \ref{correlationCoef}] as an additional metric to evaluate the linear relationship between the predicted and actual values. 

\begin{equation} \label{MAE}
\mathrm{MAE} = \frac{1}{n} \sum_{i=1}^{n} |y_i - \hat{y}_i|
\end{equation}

\begin{equation} \label{RMSE}
\mathrm{RMSE} = \sqrt{\frac{1}{n}\sum_{i=1}^{n}(y_i-\hat{y_i})^2}
\end{equation}

\begin{equation} \label{correlationCoef}
r = \frac{\sum\limits_{i=1}^{n} (y_i - \bar{y})(\hat{y_i} - \bar{\hat{y}})}{\sqrt{\sum\limits_{i=1}^{n}(y_i - \bar{y})^2 \sum\limits_{i=1}^{n}(\hat{y_i} - \bar{\hat{y}})^2}}
\end{equation}

Where n is the total number of data points, $y_{i}$ is the true value of the i-th data point, $\hat{y}_{i}$ is the predicted value of the i-th data point, and $\bar{y}$ and $\bar{\hat{y}}$ represent their respective means.

\section{Results} \label{Results}
In this section, we will first examine the results of the yield prediction, followed by the selection of the top 10 optimal genotypes that yielded the highest yields for each particular location and environmental condition.

\subsection{Prediction results} \label{Predictionresults}
The study employs the best hyperparameter settings obtained through hyperparameter tuning to train and validate three machine learning models, and our proposed hybrid deep learning models. The performance of the baseline machine learning models and the proposed GEM model in predicting soybean yield is evaluated based on the test and validation results of RMSE, MAE, and r (correlation coefficient), and it is presented in Table \ref{PrediCtionResults}. 

\begin{table}[h]
\centering
\caption{Comparison of Test and Validation Results of RMSE, MAE, and r for the Baseline Machine Learning Models and Proposed GEM Model in Predicting Soybean Yield.}
\label{PrediCtionResults}
\begin{tabular}{l c c c c}
   \hline
        \textbf{} & \textbf{Random Forest} & \textbf{XGBoost} & \textbf{Lasso regression} & \textbf{Proposed GEM} \\
        \hline
        Train RMSE & 3.24 & 6.75 & 8.69 & - \\
         Test RMSE & 7.12 & 7.04 & 9.33 & 6.67 \\
         Validation RMSE & 7.01 & 6.98 & 9.41 & 6.55 \\
         \hline
         Train MAE & 2.23 & 5.15 & 6.76 & - \\
         Test MAE & 5.32 & 5.33 & 7.26 & 5.05 \\
         Validation MAE & 5.27 & 5.32 & 7.30 & 4.92 \\
         \hline
         Train r & 0.981 & 0.908 & 0.838 & - \\
         Test r & 0.894 & 0.898 & 0.810 & 0.908 \\
         Validation r & 0.899 & 0.901 & 0.809 & 0.91 \\
        \hline
        
\end{tabular}
\end{table}

The performance of the proposed Generalized Ensemble Model (GEM), which combines the CNN-DNN model and the CNN-LSTM-DNN model, was compared with several other machine learning models including XGBoost, Random Forest, and Lasso. The results showed that the GEM model outperformed all other tested models. The reason for the outperformance can be attributed to the GEM model's ability to capture the nonlinearity of weather data and its capacity to capture the temporal dependencies of weather data.

While XGBoost and Random Forest are powerful machine learning models, they rely heavily on linear relationships and may not be able to capture the nonlinear relationships present in the weather data. Lasso, on the other hand, is a linear regression model with a L1 penalty, which can result in some of the coefficients being forced to zero. While this can result in a simpler and more interpretable model, it may not be able to capture the complex relationships present in the weather data, and G $\times$ E interactions.

The GEM model, on the other hand, combines the strengths of multiple models, including the highly nonlinear structure of the CNN model and the ability to capture the temporal dependencies of weather data using the LSTM model. This results in a more robust and accurate model that outperforms the other tested models.

\begin{figure}
\centering
\includegraphics[scale=0.50]{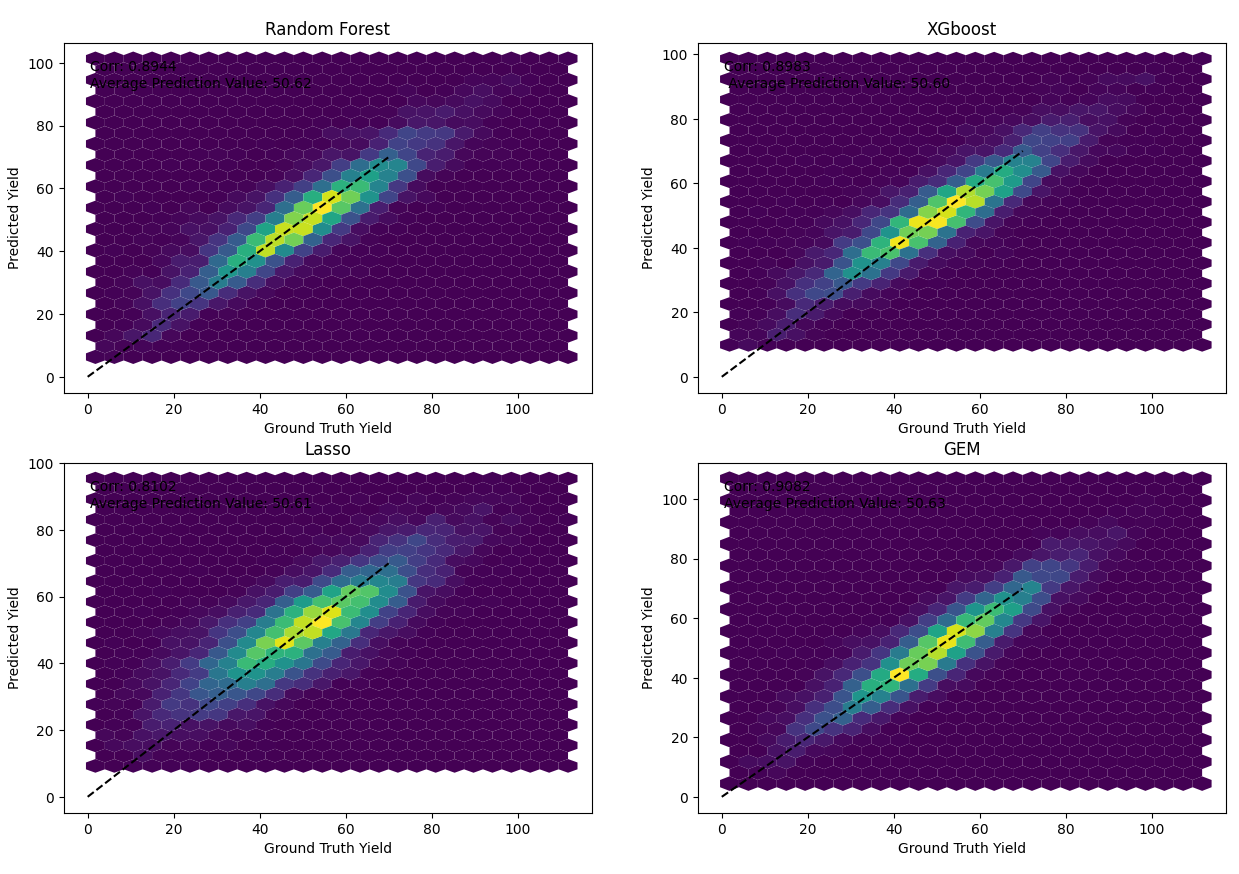}
\caption{Hexagonal plots of the predicted soybean yield vs. ground truth yield values for the three machine learning models and proposed GEM model on the test data.}\label{predictionresultsHexagonal}
\end{figure}

The hexagonal plots shown in Fig \ref{predictionresultsHexagonal} are a visualization tool used to compare the ground truth yield with the predicted yield values for different machine learning models. The plots show the density of points where the two yields overlap, with the color of the hexagons representing the density of points. The 1:1 line represents the ideal situation where the predicted yield is exactly equal to the ground truth yield.

By looking at the hexagonal plots and the position of the points relative to the 1:1 line, we can observe how well each model is performing. If the points are concentrated near the 1:1 line, it indicates that the model is performing well, with high accuracy and precision. On the other hand, if the points are scattered or far from the 1:1 line, it indicates that the model is not performing well and is making large errors in its predictions.

In this case, based on the hexagonal plots in Fig \ref{predictionresultsHexagonal}, the GEM model has the most tightly clustered predicted yield values around the 1:1 line, suggesting that it is the most accurate model in predicting soybean yield. The random forest model has a slightly wider spread of predicted yield values, indicating slightly less accuracy. The XGboost model also shows a strong positive correlation with the ground truth yield values, but has more scatter than the random forest model. The performance of the Lasso model was comparatively weaker, which was demonstrated by the scattered data points that exhibited more deviation from the 1:1 line compared to other models.

\begin{figure}
\centering
\includegraphics[scale=0.65]{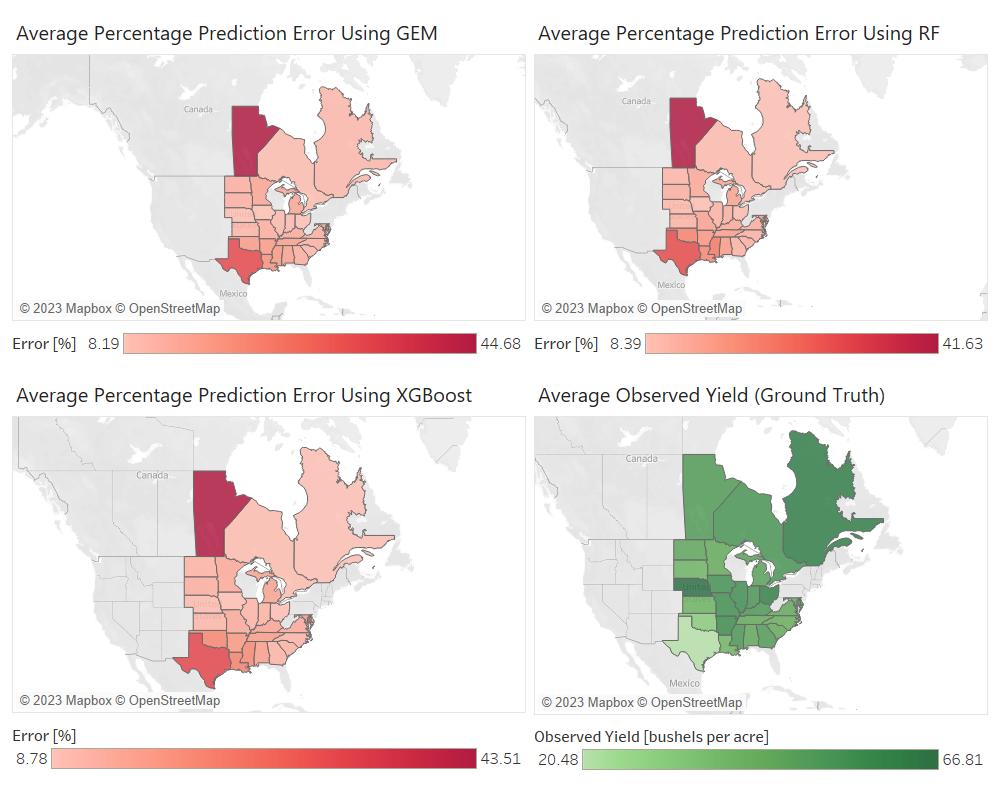}
\caption{Spatial distribution of average prediction errors for soybean yield in test data using the proposed GEM, RF, and XGBoost models, and average observed yield values across 28 U.S. states and Canadian provinces.}\label{percentagepredictionerror}
\end{figure}

Figure \ref{percentagepredictionerror} illustrates the spatial distribution of average prediction errors for soybean yield in test data using the proposed GEM, RF, and XGBoost models, and average observed yield values across 28 U.S. states and Canadian provinces. This figure allows for the identification of states/provinces with higher average error percentages, providing valuable insights to enhance data collection in those regions. The GEM model reveals a variation in average error percentages ranging from 8.19\% to 44.68\% across 28 U.S. states and Canadian provinces in the test data. As anticipated, states and provinces such as Texas and Manitoba, which have a lower number of observations as depicted in Figure \ref{datadistribution}, exhibit higher prediction errors. Although the proposed GEM model exhibits a wider range of average percentage error values compared to the RF and XGBoost models, as indicated in Table \ref{PrediCtionResults}, the overall performance of the GEM model surpasses that of the baseline models. The prediction percentage error  for location i is determined using Equation \ref{PredictionErrorPercentage}, where the difference between the predicted yield value and the actual yield value for each location within each state or province is divided by the actual yield value. We then averaged the prediction errors of all locations within each state or province and displayed the results in Figure \ref{percentagepredictionerror}.   
\begin{equation}
\text{Prediction Error Percentage}_i =  \left| \frac{\text{Actual Yield}_i - \text{Predicted Yield}_i}{\text{Actual Yield}_i} \right| \times 100 
\label{PredictionErrorPercentage}
\end{equation}

\subsection{Optimal genotype selection}

In this section, we utilized the entire dataset to identify the top 10 genotypes with the highest yields for each location-environment combination. Based on the results from the previous section, the GEM model indicated that the CNN-DNN model had the highest weight. Therefore, we proceeded to retrain the model on the entire dataset, excluding the maturity group. 
Following this, the model was used to predict yields for all 5838 genotypes across different weather and location combinations. Next, we chose the top 10 genotypes with the highest yields. We then proceeded to compute the average yield for these elite genotypes across each location-environment combination. However, due to the unavailability of weather data for all years and locations, we ended up with varying amounts of weather data for each location. For instance, for location ID 167, we selected the top 10 genotypes with the highest yields for each of weather variables from 2008 to 2010, whereas for location ID 163, we selected the top 10 genotypes with the highest yields for each weather variables from 2004 to 2012.

To illustrate the quality of these selected genotypes, we created a Tableau Public visualization. This visualization showcases the difference between the average predicted yield of optimal genotypes and the actual yield of existing genotypes across all locations in each state and year. You can access this visualization at the following link: \href{https://public.tableau.com/shared/DN4YSQR42?:display_count=n&:origin=viz_share_link}{Tableau Public Visualization}

Within this Tableau page, you have the option to select a specific year and examine the differences between the average predicted yield of optimal genotypes and the actual yield of existing genotypes across all locations in each state for that year. The range of differences observed across all years suggests that the optimal genotypes can potentially lead to increased average soybean yields in all states, with differences ranging from at least 5.1 to 42.5 bushels per acre.

\section{Analysis} \label{Analysis}
\subsection{Feature importance analysis using RMSE change}
In this study, we conducted a feature importance analysis to identify the key predictors that significantly influence our model's predictions. The analysis is based on the Root Mean Square Error (RMSE) change, which measures the impact of feature permutations on prediction performance. This method allowed us to assess the impact of variable shuffling on the model's performance.  

\textbullet\ \textbf{Baseline RMSE Calculation}: 
We initially computed the baseline RMSE (r0) using the proposed GEM model predictions (yhat) and the ground truth values (test set containing the remaining 10 \% of the data (9,303 samples)).

\textbullet\ \textbf{Permutation and RMSE Change}:
 We systematically shuffled the columns within various groups of variables and recalculated the RMSE for each permutation. These groups encompassed variables related to weather conditions, such as ADNI, AP, ARH, MDNI, MaxSur, MinSur, and AvgSur. Additionally, we considered other critical variables, including MG, year, location, and genotype ID. Among these, the categorical variables, such as MG, year, location, and genotype IDs, underwent one-hot encoding, resulting in multiple variables representing these categories. Similarly, each weather-related variable comprises 53 distinct variables, each signifying the aggregation of daily feature values through the process of averaging and downscaling the data to a 4-day granularity.

\textbullet\ \textbf{Interpreting RMSE Change}:
A higher RMSE change after shuffling indicates that the original group of variables had a more substantial impact on the model's predictions. In other words, when these variables are shuffled, the model's performance degrades significantly because they were contributing significantly to the model's accuracy.

Conversely, a lower RMSE change after shuffling suggests that the original group of variables had a lesser influence on the model's predictions. Shuffling these variables doesn't significantly impact the model's performance, indicating that they might not be as critical for prediction accuracy.

\begin{figure}
\centering
\includegraphics[scale=0.65]{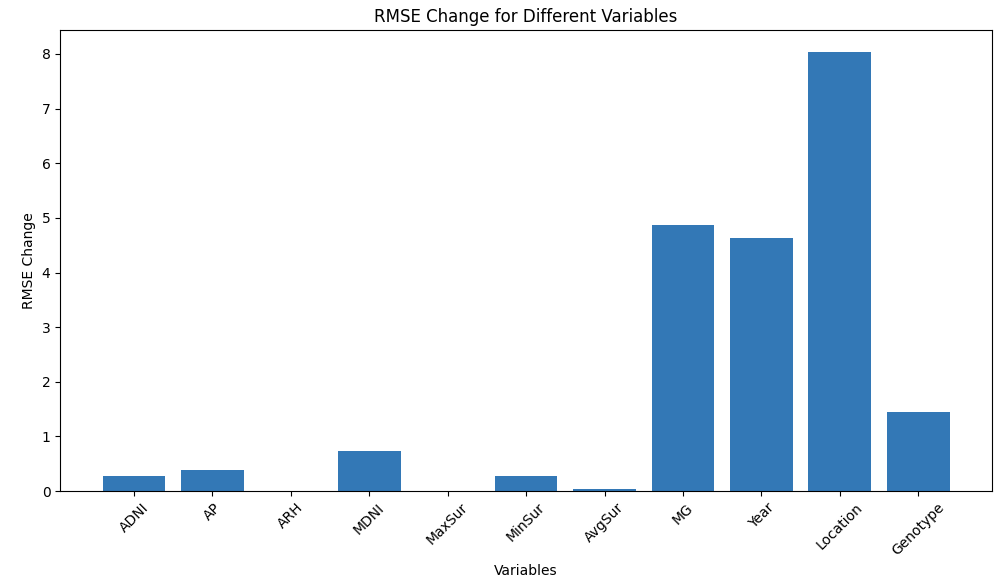}
\caption{The RMSE changes for different groups of variables after shuffling. Each group represents a set of variables, and the RMSE change quantifies the impact of shuffling those variables on the model's predictions.}\label{RMSEChangeForDifferentVariables}
\end{figure}

Fig \ref{RMSEChangeForDifferentVariables} illustrates the RMSE changes for different groups of variables after shuffling. Each group represents a set of variables, and the RMSE change quantifies the impact of shuffling those variables on the model's predictions. Each group in the analysis signifies a distinct set of variables. 

 Among all the groups, the location variable exhibits the highest RMSE change, suggesting that it plays a pivotal role in the model's predictions. When the location variable is shuffled, there is a significant decline in model performance. This emphasizes the substantial influence of geographical location on prediction accuracy. Following location, the MG variable shows the second-highest RMSE change. Shuffling this variable results in a noticeable reduction in model accuracy. This highlights the significance of considering the maturity stage of crops or plants for accurate predictions. Different maturity groups of soybeans have varying growth and flowering patterns, affecting the timing of yield. The year variable ranks third in terms of RMSE change. Shuffling the year variable leads to a substantial drop in model performance. This implies that variations across different years significantly affect the model's ability to make accurate predictions, likely due to year-specific climate patterns or other time-dependent factors. The genotype variable occupies the fourth position in RMSE change. Its shuffling causes a notable decrease in model accuracy, underscoring its importance in achieving reliable predictions. Different genotypes or plant varieties evidently contribute significantly to the model's predictive power. Within the weather category, MDNI demonstrates the highest RMSE change, followed by AP, MinSur, ADNI, AvgSur, MaxSur, and ARH. While these weather-related variables do influence the model's predictions, their impact appears to be less pronounced compared to the location, MG, year, and genotype variables. Shuffling these weather variables results in a relatively modest effect on model performance, suggesting that they may be less critical for prediction accuracy compared to the aforementioned groups.

In our analysis, we observed that certain time periods within the weather variables exhibited the highest RMSE change after shuffling. Specifically, for two key weather variables, Maximum Direct Normal Irradiance (MDNI) and Average Precipitation (AP), we identified the time periods that demonstrated the most significant impact on model performance.

\begin{figure}
\centering
\includegraphics[scale=0.65]{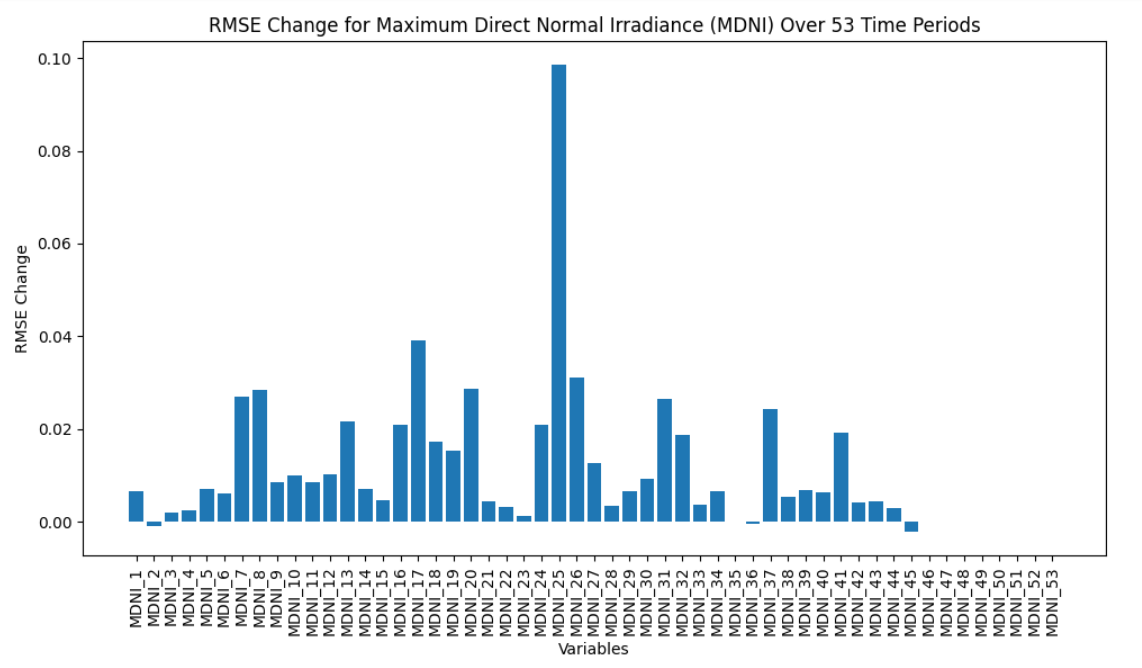}
\caption{The RMSE changes for different groups of variables after shuffling. Each group represents a set of variables, and the RMSE change quantifies the impact of shuffling those variables on the model's predictions.}\label{MDNIRMSEChange}
\end{figure}

\begin{figure}
\centering
\includegraphics[scale=0.65]{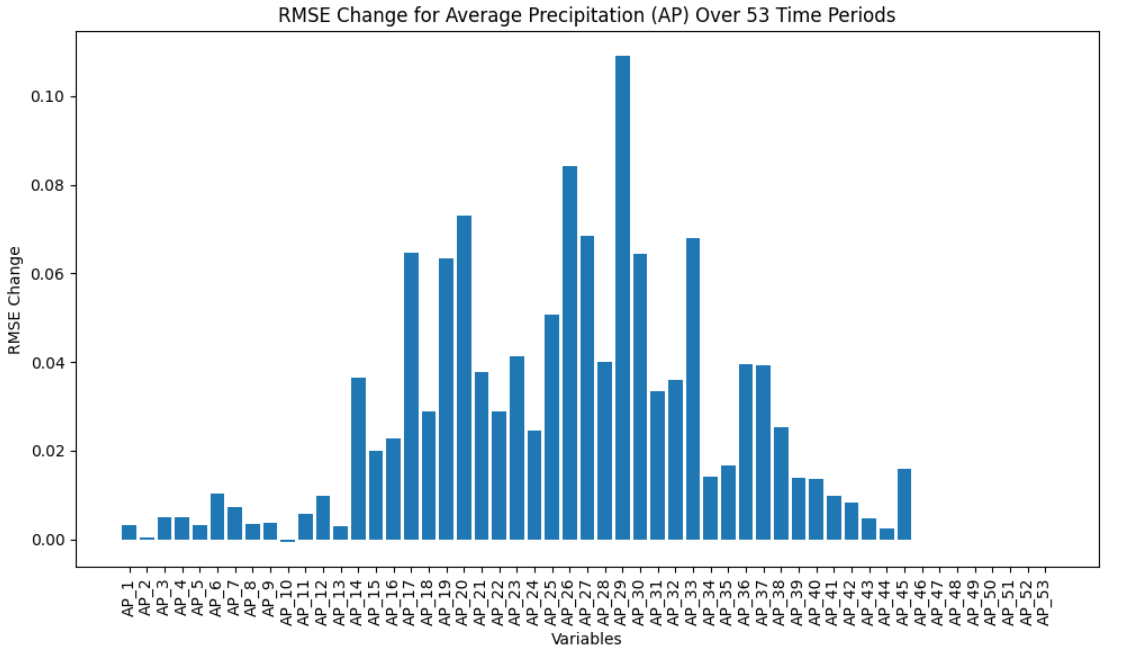}
\caption{The RMSE changes for different groups of variables after shuffling. Each group represents a set of variables, and the RMSE change quantifies the impact of shuffling those variables on the model's predictions.}\label{APRMSEChange}
\end{figure}

For MDNI, we found that time period 25, which corresponds to approximately week 15\textsuperscript{th}, exhibited the highest RMSE change following shuffling, as it is shown in Fig \ref{MDNIRMSEChange}. Similarly, for AP, time period 29 (approximately week 17\textsuperscript{th}) showed the highest RMSE change, as illustrated in Fig \ref{APRMSEChange}. These findings prompt us to explore the relationship between these time periods and the growth stages of soybeans in the United States.

In the context of soybean growth in the U.S., the growth stages are often categorized into Vegetative (V) and Reproductive (R) stages. Based on our analysis and considering typical soybean growth patterns in the USA \cite{mcwilliams1999soybean, soybeangrowthstagekentuky}, we can provide the following insights:

\textbullet\ Week 10 (Approximately): During this time, soybeans are in the early to mid-vegetative stages, typically ranging from V4 to V6. They are transitioning from early vegetative growth to the onset of reproductive growth \cite{soybeangrowthstagekentuky}.

\textbullet\ Week 12 (Approximately): At this stage (mid to late June), soybeans are typically in the V6 to V8 vegetative stage, indicating that they are approaching the reproductive stages \cite{soybeangrowthstagekentuky}.

\textbullet\ Week 15 (Approximately): This period, occurring in early to mid-July, corresponds to soybeans being in the V8 to V10 vegetative stage. This is a critical time when soybeans start transitioning to early reproductive stages, with some plants beginning to flower (R1 stage) \cite{soybeangrowthstagekentuky}.

\textbullet\ Week 17 (Approximately): Around mid to late July, soybeans may have progressed to the R2 (Full Flower) to R3 (Beginning Pod) stages. This is a vital phase during which soybeans flower and initiate pod development \cite{soybeangrowthstagekentuky}.

The observed highest RMSE changes in time periods 25 (MDNI) and 29 (AP) suggest a noteworthy correlation with soybean growth stages. The RMSE changes in these weeks signify the sensitivity of soybean growth to solar radiation (MDNI) and precipitation (AP) during key reproductive and pod development stages. These findings underscore the significance of weather variables during crucial growth phases of soybeans and their influence on accurate yield predictions.

\section{Conclusion} \label{Conclusion}

In this study, we proposed two novel convolutional neural network (CNN) architectures that incorporate a 1-D convolution operation and a long short-term memory (LSTM) layer. These models were developed to predict soybean yield using a combination of factors, including maturity group (MG), genotype ID, year, location, and weather data. Our study is based on an extensive dataset collected from 159 locations across 28 U.S. states and Canadian provinces over a span of 13 years. These architectures represent a significant advancement in the field of crop yield prediction, allowing us to leverage the power of deep learning to improve accuracy and efficiency in genotype selection. Moreover, we have employed the Generalized Ensemble Method (GEM) to determine the optimal weights of our proposed CNN-based models, which has led to superior performance in the MLCAS2021 Crop Yield Prediction Challenge and compared to baseline models. 

Our work has gone beyond traditional crop yield prediction methods by addressing the challenge of genotype × environment interaction, which is a critical factor in selecting genotypes for increased crop yields, particularly in the face of global climate change. Conventionally, plant breeders rely on extensive field testing of hybrids to identify those with the highest yield potential, a process that is both time-consuming and resource-intensive. Our approach has introduced a data-driven paradigm for genotype selection, wherein we use environmental data and genotype information to predict crop yields. This approach enables us to identify the most efficient genotypes for each location and environmental condition by forecasting crop yields based on weather conditions and then selecting the optimal genotype with the highest yield. This novel strategy holds the potential to significantly enhance policy and agricultural decision-making, optimize production, and ensure food security.

In our analysis, we evaluated our proposed GEM model against three commonly used prediction models: random forest (RF), extreme gradient boosting (XGBoost), and least absolute shrinkage and selection operator (LASSO). The results unequivocally demonstrated the superiority of the GEM model. The GEM model's exceptional performance can be attributed to its ability to capture the nonlinear nature of weather data and its capacity to model the temporal dependencies of weather variables, including genotype $\times$ environment interactions. This is made possible by combining two CNN-based models, which are particularly good at handling these complex relationships in the data.

Additionally, we conducted a feature importance analysis using RMSE change to identify significant predictors affecting the model's predictions. The location variable had the highest RMSE change, indicating its strong influence on predictions. MG, year, and genotype also played crucial roles. Among weather variables, Maximum Direct Normal Irradiance (MDNI) had the most impact, followed by Average Precipitation (AP), Minimum Surface Temperature (MinSur), and others. While weather variables influenced predictions, categorical variables like location and MG were more influential.

In addition to assessing the importance of different variable groups, we delved deeper into the temporal aspects of weather data. Specifically, we investigated significant time periods within MDNI and AP variables that exhibited the highest RMSE change after shuffling. The highest RMSE changes observed in time periods 25 (week 15\textsuperscript{th}) (MDNI), and 29 (week 17\textsuperscript{th}) (AP) point to a significant link with soybean growth stages. These RMSE fluctuations in these weeks highlight how soybean growth is affected by solar radiation (MDNI) and precipitation (AP) during important reproductive and pod development stages. 

Because the climate change can also have an adverse effects on soil attributes \cite{das2016potential}, it is suggested to consider a detaset which also includes the soil variables to better understand the complex interplay between weather, soil, and crop outcomes. Further research could explore the inclusion of soil attributes in the predictive model to enhance accuracy, especially in regions where soil quality significantly impacts agricultural outcomes. Additionally, investigating interactions between weather, soil, and crop variables could lead to more precise and robust predictive models, ultimately benefiting agriculture and food security in the face of changing climate conditions.

\section{Acknowledgments} \label{Acknowledgments}
We express our gratitude to the organizers of the Third International Workshop on Machine Learning for Cyber-Agricultural Systems (MLCAS 2021) for their diligent efforts in arranging the MLCAS2021 Crop Yield Prediction Challenge and generously sharing the invaluable datasets. It is noteworthy that this manuscript has also been made available as a preprint on arXiv.





\bibliographystyle{unsrtnat}
\bibliography{references}  

\begin{thebibliography}{44}
\providecommand{\natexlab}[1]{#1}
\providecommand{\url}[1]{\texttt{#1}}
\expandafter\ifx\csname urlstyle\endcsname\relax
  \providecommand{\doi}[1]{doi: #1}\else
  \providecommand{\doi}{doi: \begingroup \urlstyle{rm}\Url}\fi

\bibitem[Nations et~al.(2017)Nations, of~Economic, and
  Social~Affairs]{united2017world}
United Nations, Department of~Economic, and Population~Division Social~Affairs.
\newblock World population prospects: the 2017 revision, key findings and
  advance tables.
\newblock \emph{Working Paper No. ESA/P/WP/248 ed}, 2017.

\bibitem[Kumar(2016)]{kumar2016impact}
Manoj Kumar.
\newblock Impact of climate change on crop yield and role of model for
  achieving food security.
\newblock \emph{Environmental Monitoring and Assessment}, 188\penalty0
  (8):\penalty0 1--14, 2016.

\bibitem[Xiong et~al.(2022)Xiong, Reynolds, and Xu]{xiong2022climate}
Wei Xiong, Matthew Reynolds, and Yunbi Xu.
\newblock Climate change challenges plant breeding.
\newblock \emph{Current Opinion in Plant Biology}, 70:\penalty0 102308, 2022.

\bibitem[de~Los~Campos et~al.(2020)de~Los~Campos, P{\'e}rez-Rodr{\'\i}guez,
  Bogard, Gouache, and Crossa]{de2020data}
Gustavo de~Los~Campos, Paulino P{\'e}rez-Rodr{\'\i}guez, Matthieu Bogard, David
  Gouache, and Jos{\'e} Crossa.
\newblock A data-driven simulation platform to predict cultivars’
  performances under uncertain weather conditions.
\newblock \emph{Nature communications}, 11\penalty0 (1):\penalty0 1--10, 2020.

\bibitem[Heslot et~al.(2014)Heslot, Akdemir, Sorrells, and
  Jannink]{heslot2014integrating}
Nicolas Heslot, Deniz Akdemir, Mark~E Sorrells, and Jean-Luc Jannink.
\newblock Integrating environmental covariates and crop modeling into the
  genomic selection framework to predict genotype by environment interactions.
\newblock \emph{Theoretical and applied genetics}, 127\penalty0 (2):\penalty0
  463--480, 2014.

\bibitem[Roberts et~al.(2017)Roberts, Braun, Sinclair, Lobell, and
  Schlenker]{roberts2017comparing}
Michael~J Roberts, Noah~O Braun, Thomas~R Sinclair, David~B Lobell, and Wolfram
  Schlenker.
\newblock Comparing and combining process-based crop models and statistical
  models with some implications for climate change.
\newblock \emph{Environmental Research Letters}, 12\penalty0 (9):\penalty0
  095010, 2017.

\bibitem[Hajjarpoor et~al.(2022)Hajjarpoor, Nelson, and
  Vadez]{hajjarpoor2022process}
Amir Hajjarpoor, William~CD Nelson, and Vincent Vadez.
\newblock How process-based modeling can help plant breeding deal with g x e x
  m interactions.
\newblock \emph{Field Crops Research}, 283:\penalty0 108554, 2022.

\bibitem[Shahhosseini et~al.(2021)Shahhosseini, Hu, Khaki, and
  Archontoulis]{shahhosseini2021corn}
Mohsen Shahhosseini, Guiping Hu, Saeed Khaki, and Sotirios~V Archontoulis.
\newblock Corn yield prediction with ensemble cnn-dnn.
\newblock \emph{Frontiers in plant science}, 12:\penalty0 709008, 2021.

\bibitem[Khaki and Wang(2019)]{khaki2019crop}
Saeed Khaki and Lizhi Wang.
\newblock Crop yield prediction using deep neural networks.
\newblock \emph{Frontiers in plant science}, 10:\penalty0 621, 2019.

\bibitem[Crane-Droesch(2018)]{crane2018machine}
Andrew Crane-Droesch.
\newblock Machine learning methods for crop yield prediction and climate change
  impact assessment in agriculture.
\newblock \emph{Environmental Research Letters}, 13\penalty0 (11):\penalty0
  114003, 2018.

\bibitem[Xu et~al.(2022)Xu, Zhang, Li, Zheng, Zhang, Olsen, Varshney, Prasanna,
  and Qian]{xu2022smart}
Yunbi Xu, Xingping Zhang, Huihui Li, Hongjian Zheng, Jianan Zhang, Michael~S
  Olsen, Rajeev~K Varshney, Boddupalli~M Prasanna, and Qian Qian.
\newblock Smart breeding driven by big data, artificial intelligence and
  integrated genomic-enviromic prediction.
\newblock \emph{Molecular Plant}, 2022.

\bibitem[Patil and Thorat(2016)]{patil2016early}
Suyash~S Patil and Sandeep~A Thorat.
\newblock Early detection of grapes diseases using machine learning and iot.
\newblock In \emph{2016 second international conference on Cognitive Computing
  and Information Processing (CCIP)}, pages 1--5. IEEE, 2016.

\bibitem[Chen et~al.(2021)Chen, Goh, Sin, Lim, Chung, and
  Liew]{chen2021automated}
Zhiyuan Chen, Howe~Seng Goh, Kai~Ling Sin, Kelly Lim, Nicole Ka~Hei Chung, and
  Xin~Yu Liew.
\newblock Automated agriculture commodity price prediction system with machine
  learning techniques.
\newblock \emph{arXiv preprint arXiv:2106.12747}, 2021.

\bibitem[Lowe et~al.(2022)Lowe, Qin, and Mao]{lowe2022review}
Matthew Lowe, Ruwen Qin, and Xinwei Mao.
\newblock A review on machine learning, artificial intelligence, and smart
  technology in water treatment and monitoring.
\newblock \emph{Water}, 14\penalty0 (9):\penalty0 1384, 2022.

\bibitem[Domingues et~al.(2022)Domingues, Brand{\~a}o, and
  Ferreira]{domingues2022machine}
Tiago Domingues, Tom{\'a}s Brand{\~a}o, and Jo{\~a}o~C Ferreira.
\newblock Machine learning for detection and prediction of crop diseases and
  pests: A comprehensive survey.
\newblock \emph{Agriculture}, 12\penalty0 (9):\penalty0 1350, 2022.

\bibitem[Sharma et~al.(2020)Sharma, Jain, Gupta, and
  Chowdary]{sharma2020machine}
Abhinav Sharma, Arpit Jain, Prateek Gupta, and Vinay Chowdary.
\newblock Machine learning applications for precision agriculture: A
  comprehensive review.
\newblock \emph{IEEE Access}, 9:\penalty0 4843--4873, 2020.

\bibitem[Bertan et~al.(2007)Bertan, de~Carvalho, and
  Oliveira]{bertan2007parental}
Ivandro Bertan, Fernando~IF de~Carvalho, and Antonio Costa~de Oliveira.
\newblock Parental selection strategies in plant breeding programs.
\newblock \emph{Journal of crop science and biotechnology}, 10\penalty0
  (4):\penalty0 211--222, 2007.

\bibitem[Khaki et~al.(2020{\natexlab{a}})Khaki, Khalilzadeh, and
  Wang]{khaki2020predicting}
Saeed Khaki, Zahra Khalilzadeh, and Lizhi Wang.
\newblock Predicting yield performance of parents in plant breeding: A neural
  collaborative filtering approach.
\newblock \emph{Plos one}, 15\penalty0 (5):\penalty0 e0233382,
  2020{\natexlab{a}}.

\bibitem[Arzanipour and Olafsson(2022)]{arzanipour2022evaluating}
Atousa Arzanipour and Sigurdur Olafsson.
\newblock Evaluating imputation in a two-way table of means for training data
  construction.
\newblock \emph{SSRN}, 2022.
\newblock \doi{10.2139/ssrn.4476272}.
\newblock URL \url{https://ssrn.com/abstract=4476272}.

\bibitem[Srivastava et~al.(2022)Srivastava, Safaei, Khaki, Lopez, Zeng, Ewert,
  Gaiser, and Rahimi]{srivastava2022winter}
Amit~Kumar Srivastava, Nima Safaei, Saeed Khaki, Gina Lopez, Wenzhi Zeng, Frank
  Ewert, Thomas Gaiser, and Jaber Rahimi.
\newblock Winter wheat yield prediction using convolutional neural networks
  from environmental and phenological data.
\newblock \emph{Scientific Reports}, 12\penalty0 (1):\penalty0 3215, 2022.

\bibitem[Shook et~al.(2021)Shook, Gangopadhyay, Wu, Ganapathysubramanian,
  Sarkar, and Singh]{shook2021crop}
Johnathon Shook, Tryambak Gangopadhyay, Linjiang Wu, Baskar
  Ganapathysubramanian, Soumik Sarkar, and Asheesh~K Singh.
\newblock Crop yield prediction integrating genotype and weather variables
  using deep learning.
\newblock \emph{Plos one}, 16\penalty0 (6):\penalty0 e0252402, 2021.

\bibitem[Veenadhari et~al.(2011)Veenadhari, Mishra, and
  Singh]{veenadhari2011soybean}
S~Veenadhari, Bharat Mishra, and CD~Singh.
\newblock Soybean productivity modelling using decision tree algorithms.
\newblock \emph{International Journal of Computer Applications}, 27\penalty0
  (7):\penalty0 11--15, 2011.

\bibitem[Kang et~al.(2020)Kang, Ozdogan, Zhu, Ye, Hain, and
  Anderson]{kang2020comparative}
Yanghui Kang, Mutlu Ozdogan, Xiaojin Zhu, Zhiwei Ye, Christopher Hain, and
  Martha Anderson.
\newblock Comparative assessment of environmental variables and machine
  learning algorithms for maize yield prediction in the us midwest.
\newblock \emph{Environmental Research Letters}, 15\penalty0 (6):\penalty0
  064005, 2020.

\bibitem[Dorffner(1996)]{dorffner1996neural}
Georg Dorffner.
\newblock Neural networks for time series processing.
\newblock \emph{Neural network world}, 6\penalty0 (4):\penalty0 447--468, 1996.

\bibitem[Albawi et~al.(2017)Albawi, Mohammed, and
  Al-Zawi]{albawi2017understanding}
Saad Albawi, Tareq~Abed Mohammed, and Saad Al-Zawi.
\newblock Understanding of a convolutional neural network.
\newblock In \emph{2017 international conference on engineering and technology
  (ICET)}, pages 1--6. Ieee, 2017.

\bibitem[Sherstinsky(2020)]{sherstinsky2020fundamentals}
Alex Sherstinsky.
\newblock Fundamentals of recurrent neural network (rnn) and long short-term
  memory (lstm) network.
\newblock \emph{Physica D: Nonlinear Phenomena}, 404:\penalty0 132306, 2020.

\bibitem[Lipton et~al.(2015)Lipton, Berkowitz, and Elkan]{lipton2015critical}
Zachary~C Lipton, John Berkowitz, and Charles Elkan.
\newblock A critical review of recurrent neural networks for sequence learning.
\newblock \emph{arXiv preprint arXiv:1506.00019}, 2015.

\bibitem[Hochreiter and Schmidhuber(1997)]{hochreiter1997long}
Sepp Hochreiter and J{\"u}rgen Schmidhuber.
\newblock Long short-term memory.
\newblock \emph{Neural computation}, 9\penalty0 (8):\penalty0 1735--1780, 1997.

\bibitem[Hochreiter and Schmidhuber(1996)]{hochreiter1996lstm}
Sepp Hochreiter and J{\"u}rgen Schmidhuber.
\newblock Lstm can solve hard long time lag problems.
\newblock \emph{Advances in neural information processing systems}, 9, 1996.

\bibitem[Xie et~al.(2019)Xie, Liang, Liang, and Zhao]{xie2019attention}
Yue Xie, Ruiyu Liang, Zhenlin Liang, and Li~Zhao.
\newblock Attention-based dense lstm for speech emotion recognition.
\newblock \emph{IEICE TRANSACTIONS on Information and Systems}, 102\penalty0
  (7):\penalty0 1426--1429, 2019.

\bibitem[Li et~al.(2019)Li, Zhou, Chen, and Gao]{li2019residual}
Xiangpeng Li, Zhilong Zhou, Lijiang Chen, and Lianli Gao.
\newblock Residual attention-based lstm for video captioning.
\newblock \emph{World Wide Web}, 22:\penalty0 621--636, 2019.

\bibitem[Sun et~al.(2019)Sun, Di, Sun, Shen, and Lai]{sun2019county}
Jie Sun, Liping Di, Ziheng Sun, Yonglin Shen, and Zulong Lai.
\newblock County-level soybean yield prediction using deep cnn-lstm model.
\newblock \emph{Sensors}, 19\penalty0 (20):\penalty0 4363, 2019.

\bibitem[Gangopadhyay et~al.(2018)Gangopadhyay, Tan, Huang, and
  Sarkar]{gangopadhyay2018temporal}
Tryambak Gangopadhyay, Sin~Yong Tan, Genyi Huang, and Soumik Sarkar.
\newblock Temporal attention and stacked lstms for multivariate time series
  prediction.
\newblock 2018.

\bibitem[Sutskever et~al.(2014)Sutskever, Vinyals, and
  Le]{sutskever2014sequence}
Ilya Sutskever, Oriol Vinyals, and Quoc~V Le.
\newblock Sequence to sequence learning with neural networks.
\newblock \emph{Advances in neural information processing systems}, 27, 2014.

\bibitem[Khaki et~al.(2020{\natexlab{b}})Khaki, Wang, and
  Archontoulis]{khaki2020cnn}
Saeed Khaki, Lizhi Wang, and Sotirios~V Archontoulis.
\newblock A cnn-rnn framework for crop yield prediction.
\newblock \emph{Frontiers in Plant Science}, 10:\penalty0 1750,
  2020{\natexlab{b}}.

\bibitem[Oikonomidis et~al.(2022)Oikonomidis, Catal, and
  Kassahun]{oikonomidis2022hybrid}
Alexandros Oikonomidis, Cagatay Catal, and Ayalew Kassahun.
\newblock Hybrid deep learning-based models for crop yield prediction.
\newblock \emph{Applied artificial intelligence}, 36\penalty0 (1):\penalty0
  2031822, 2022.

\bibitem[{MLCAS, 2021}()]{mlcas2021challenge}
{MLCAS, 2021}.
\newblock {MLCAS2021 Crop Yield Prediction Challenge}.
\newblock URL
  \url{https://eval.ai/web/challenges/challenge-page/1251/overview}.

\bibitem[Perrone and Cooper(1995)]{perrone1995networks}
Michael~P Perrone and Leon~N Cooper.
\newblock When networks disagree: Ensemble methods for hybrid neural networks.
\newblock In \emph{How We Learn; How We Remember: Toward An Understanding Of
  Brain And Neural Systems: Selected Papers of Leon N Cooper}, pages 342--358.
  World Scientific, 1995.

\bibitem[Breiman(2001)]{breiman2001random}
Leo Breiman.
\newblock Random forests.
\newblock \emph{Machine learning}, 45:\penalty0 5--32, 2001.

\bibitem[Chen and Guestrin(2016)]{chen2016xgboost}
Tianqi Chen and Carlos Guestrin.
\newblock Xgboost: A scalable tree boosting system.
\newblock In \emph{Proceedings of the 22nd acm sigkdd international conference
  on knowledge discovery and data mining}, pages 785--794, 2016.

\bibitem[Tibshirani(1996)]{tibshirani1996regression}
Robert Tibshirani.
\newblock Regression shrinkage and selection via the lasso.
\newblock \emph{Journal of the Royal Statistical Society: Series B
  (Methodological)}, 58\penalty0 (1):\penalty0 267--288, 1996.

\bibitem[McWilliams et~al.(1999)McWilliams, Berglund, and
  Endres]{mcwilliams1999soybean}
Denise~A McWilliams, Duane~Raymond Berglund, and GJ~Endres.
\newblock Soybean growth and management quick guide.
\newblock 1999.

\bibitem[{University of Kentucky Cooperative
  Extension}(n.d.)]{soybeangrowthstagekentuky}
{University of Kentucky Cooperative Extension}.
\newblock Soybean growth and development, n.d.
\newblock URL \url{https://www2.ca.uky.edu/agcomm/pubs/AGR/AGR223/AGR223.pdf}.
\newblock Accessed: 9/13/2023.

\bibitem[Das et~al.(2016)Das, Dutta, and Rakshit]{das2016potential}
Indranil Das, Debashis Dutta, and Amitava Rakshit.
\newblock Potential effects of climate change on soil properties: A review.
\newblock 2016.

\end{thebibliography}






\end{document}